\documentclass[10pt,twocolumn,letterpaper]{article}

\usepackage{cvpr}              

\newcommand{\red}[1]{{\color{red}#1}}

\definecolor{cvprblue}{rgb}{0.21,0.49,0.74}
\usepackage[pagebackref,breaklinks,colorlinks,allcolors=cvprblue]{hyperref}

\usepackage{amsmath,amssymb,amsfonts}
\usepackage{algorithmic}
\usepackage{graphicx,color}
\usepackage{textcomp}
\usepackage{xcolor}
\usepackage{hyperref}
\hypersetup{hidelinks}
\usepackage{algorithm,algorithmic}
\usepackage{adjustbox}
\usepackage{diagbox}

\usepackage{multirow}
\usepackage{hyperref}
\newcommand{\bm}[1]{\mbox{\boldmath ${#1}$}}
\newcommand{\blue}[1]{\bf{\textcolor{blue}{#1}}}

\usepackage[accsupp]{axessibility}  

\def\paperID{43} 
\def\confName{CVPR}
\def\confYear{2025}

\title{Efficient Burst Super-Resolution with One-step Diffusion}

\author{Kento Kawai ~~~ Takeru Oba ~~~ Kyotaro Tokoro ~~~ Kazutoshi Akita ~~~ Norimichi Ukita\\
Toyota Technological Institute\\
{\tt\small \{sd21033,sd21502,sd24439,sd21501,ukita\}@toyota-ti.ac.jp}
}

\begin{document}
\maketitle

\begin{abstract}
While burst Low-Resolution  (LR) images are useful for improving their Super Resolution (SR) image
compared to a single LR image,
prior burst SR methods are trained in a deterministic manner, which produces a blurry SR image.
Since such blurry images are perceptually degraded, we aim to reconstruct sharp and high-fidelity SR images by a diffusion model.
Our method improves the efficiency of the diffusion model with a stochastic sampler with a high-order ODE as well as one-step diffusion using knowledge distillation.
Our experimental results demonstrate that our method can reduce the runtime to 1.6 \% of its baseline while maintaining the SR quality measured based on image distortion and perceptual quality.
\end{abstract}


\section{Introduction}

Super-Resolution  (SR) is a task for super-resolving Low-Resolution  (LR) images to their High-Resolution  (HR) images.
Among various SR methods, SR for super-resolving an LR image is called Single-Image Super-Resolution  (SISR)~\cite{DBLP:journals/pami/DongLHT16,ntire2018,DBLP:conf/cvpr/HarisSU18,DBLP:conf/iccvw/GuLZXYZYSTDLDLG19,DBLP:journals/pami/HarisSU21}.
However, SISR is not an easy task due to its ill-posed nature. That is, there are multiple appropriate HR images of each LR image.
In addition to this ill-posed nature, various image degradations make SISR more difficult.

\begin{figure}[t]
  \begin{center}
  \includegraphics[width=\linewidth]{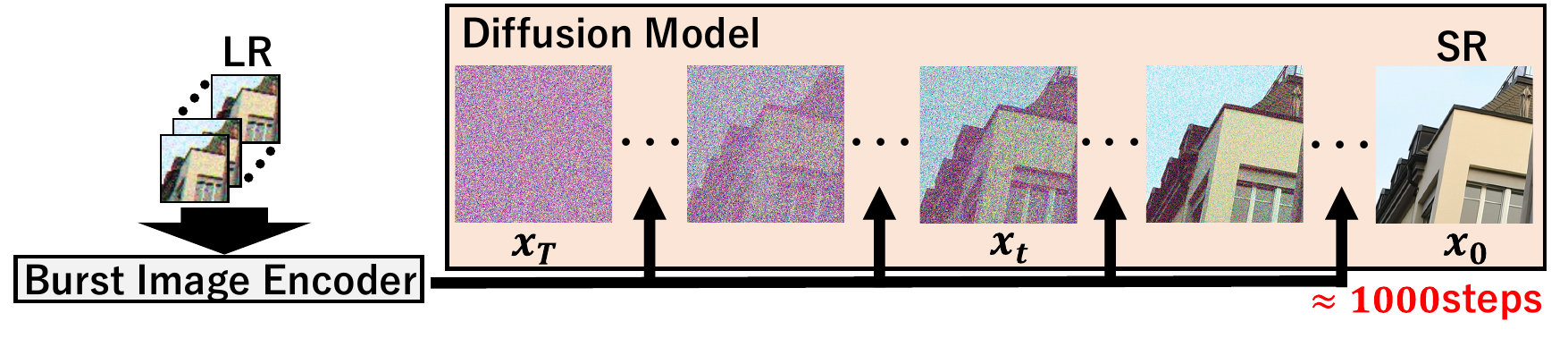}\\
   (a) SR using diffusion from random noise\\
  \vspace*{2mm}
  \includegraphics[width=\linewidth]{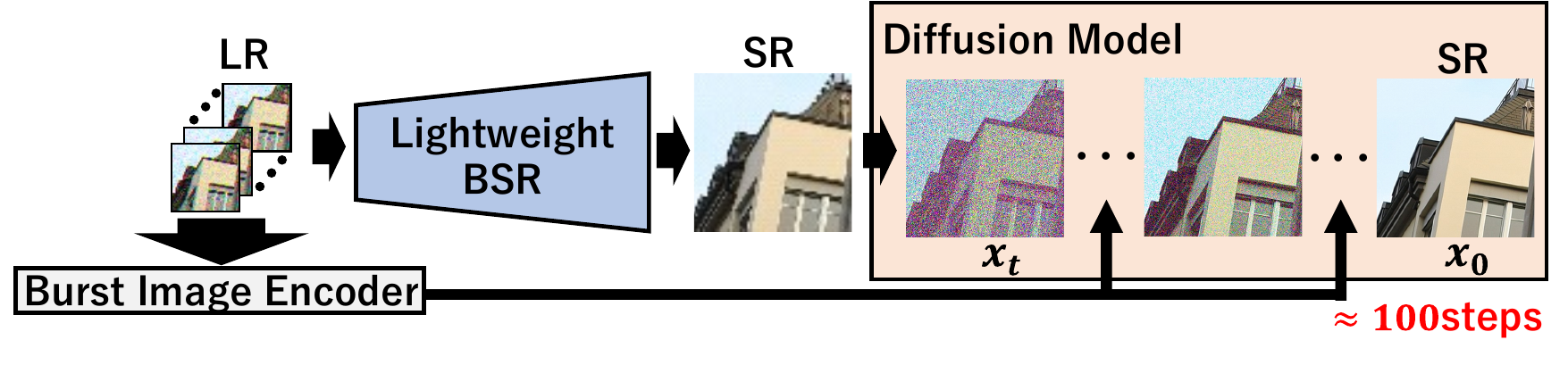}\\
   (b) BSRD~\cite{DBLP:conf/ijcnn/TokoroAU24}\\
  \vspace*{2mm}
  \includegraphics[width=\linewidth]{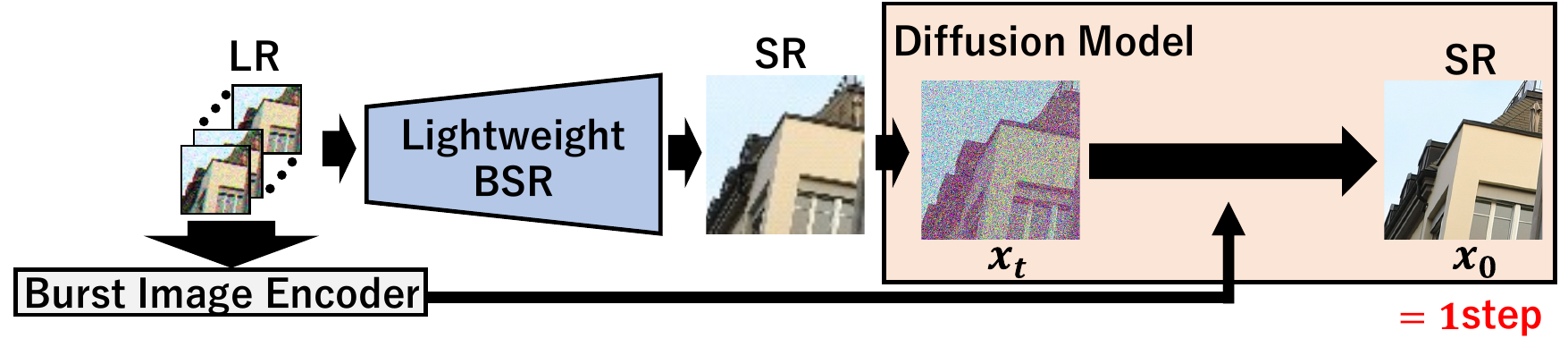}\\
   (c) Our E-BSRD  (Sec.~\ref{section: one-step})
  \caption{Comparison between prior SR model and our burst SR models.}
  \label{fig: teaser}
  \end{center}
  \vspace*{-4mm}
\end{figure}

To improve the SR quality, SR methods can take multiple differently-degraded LR images instead of a single LR image to compensate for the degradations between these LR images.
Such differently-degraded images can be easily captured by the burst shot mode of a smartphone.
Burst SR~\cite{DBLP:journals/corr/abs-2106-03839,DBLP:conf/cvpr/BhatDTCCCCCDFGG22,DBLP:conf/cvpr/BhatDGT21,DBLP:conf/cvpr/LuoYMLJF0L21,DBLP:conf/cvpr/Guo0MR022,DBLP:conf/cvpr/LuoLCYWWFSL22,DBLP:conf/cvpr/DudhaneZ0K022,DBLP:conf/cvpr/MehtaDMZ0K22,DBLP:conf/cvpr/DudhaneZ0K023} integrates these burst images into a single SR image.

Previous burst SR methods~\cite{DBLP:conf/cvpr/BhatDGT21,DBLP:conf/cvpr/LuoYMLJF0L21,DBLP:conf/cvpr/Guo0MR022,DBLP:conf/cvpr/LuoLCYWWFSL22,DBLP:conf/cvpr/DudhaneZ0K022,DBLP:conf/cvpr/MehtaDMZ0K22,DBLP:conf/cvpr/DudhaneZ0K023} are trained in a deterministic manner
based mainly on a difference between the SR image and its ground truth HR image, such as L1 and L2 losses.
However, such reconstruction losses lead to blurred SR images because these losses can be minimized by the average of HR images~\cite{DBLP:conf/cvpr/LedigTHCCAATTWS17}, each of which is an appropriate HR image of the input LR image.

To avoid this problem, this paper aims to improve the burst SR quality by probabilistic modeling.
Probabilistic modeling allows us to represent the probabilistic distribution of sharp SR images of each LR image.
Among such probabilistic models, diffusion models~\cite{DBLP:conf/nips/HoJA20} are used by the recent burst SR method~\cite{DBLP:conf/ijcnn/TokoroAU24}, as shown in  (Fig.~\ref{fig: teaser}  (b)), because of their performance validated in various computer vision tasks.

To employ the diffusion models for burst SR, this paper focuses on the following two issues:

\noindent
{\bf How to efficiently achieve burst SR.}
In diffusion models proposed earlier~\cite{DBLP:conf/nips/HoJA20}, a huge number of iterations  (e.g., around 1,000) are required to generate 
realistic data such as natural images.
Since such a huge number of iterations are computationally expensive,
several approaches are proposed to reduce the iterative steps  (e.g., model minimization using distillation~\cite{DBLP:conf/icml/SongD0S23,DBLP:conf/iclr/KimLLMTUHME24}
and efficient sampling and training processes~\cite{DBLP:conf/nips/KarrasAAL22}).
While these approaches are applied to SISR~\cite{DBLP:conf/cvpr/ChungSY22, DBLP:conf/cvpr/WangYCWGCLQKW24,DBLP:conf/nips/YueWL23}, they are not optimized for burst SR.

\noindent
{\bf How to integrate differently-degraded LR images.}
Unlike random image generation~\cite{DBLP:conf/nips/HoJA20}, image enhancement and restoration, including SR, must condition a diffusion model for generating an image that fits with input images.
For example, for diffusion-based SISR~\cite{DBLP:journals/pami/SahariaHCSFN23,DBLP:journals/corr/abs-2305-07015}, an LR image is used for conditioning the reverse diffusion process.
Unlike SISR, however, burst SR takes multiple LR images with different degradations, including blur, noise, and displacement.
If such differently-degraded images are directly fed into the diffusion model, the reverse process may reconstruct a blurry SR image
by averaging these degraded images.

To cope with these two issues, CCDF~\cite{DBLP:conf/cvpr/ChungSY22} and BSRD~\cite{DBLP:conf/ijcnn/TokoroAU24} efficiently uses a diffusion model by skipping early diffusion steps.
This skip is achieved by feeding an initial burst SR image reconstructed by a simple deterministic burst SR method into the intermediate step of a probabilistic diffusion model, as shown in Fig.~\ref{fig: teaser}  (b).
Furthermore, unlike previous diffusion-based SISR methods~\cite{DBLP:conf/cvpr/ChungSY22, DBLP:conf/cvpr/WangYCWGCLQKW24,DBLP:conf/nips/YueWL23}, BSRD employs spatially-aligned multi-scale features extracted from input burst LR images for conditioning the diffusion model.

Our paper extends
BSRD~\cite{DBLP:conf/ijcnn/TokoroAU24} by incorporating the following advancements:
\begin{itemize}
\item Instead of diffusion models using first-order differential equations used in~\cite{DBLP:conf/cvpr/ChungSY22} and \cite{DBLP:conf/ijcnn/TokoroAU24}, such as DDPM~\cite{DBLP:conf/nips/HoJA20}, DDIM~\cite{DBLP:conf/iclr/SongME21}, iDDPM  (variance scheduling)~\cite{DBLP:conf/icml/NicholD21}, and VP/VE~\cite{DBLP:conf/iclr/0011SKKEP21}, a second-order differential equation-based model  (i.e., Elucidating the Design Space of Diffusion-Based
Generative Model  (EDM)~\cite{DBLP:conf/nips/KarrasAAL22}) is used to reduce the number of diffusion steps as well as to improve the SR quality.
While the original EDM removes a large amount of noise per step from random noise to reduce the number of diffusion steps, each step in our method removes only a small amount of noise by appropriately decreasing the noise given to an initial SR image, enabling fine-grained SR reconstruction even through a smaller number of diffusion steps.
\item For further reducing the number of diffusion steps for efficient burst SR  (i.e., between 5 and 100 in~\cite{DBLP:conf/cvpr/ChungSY22} and \cite{DBLP:conf/ijcnn/TokoroAU24} to
one step in our method),
a distillation-based teacher-student model~\cite{DBLP:conf/icml/SongD0S23,DBLP:conf/iclr/KimLLMTUHME24} is employed.
Since this distillation-based model degrades the quality of image synthesis if it begins from random noise,
our method avoids this degradation by feeding a properly initialized SR image, which is reconstructed by simple deterministic Burst SR in our method, into the diffusion model.
\item  Several important parameters for EDM and the distillation model are optimized for our burst SR model.
\item With the aforementioned contributions, the SR reconstruction time is reduced to 1.6 \% of the baseline, i.e., BSRD~\cite{DBLP:conf/ijcnn/TokoroAU24}.
\end{itemize}


\section{Related Work}
\label{section: related}

\subsection{Burst SR}
\label{subsection: related_burst_SR}

While a RAW image consisting of RGGB channels  (4 channels) with high bits per pixel resolution  (e.g., 14 bits, 16 bits) is captured by a standard digital camera, it is converted
by an Image Signal Processing  (ISP)
to its 8-bit RGB image.
General burst SR methods take a set of unprocessed RAW images instead of their processed RGB images.
Burst SR methods generally consist of four processes, i.e., feature extraction, alignment, fusion, and reconstruction processes.
The alignment process rectifies image features extracted by the feature extraction process so that the features are spatially consistent.
The aligned features of all burst images are then merged by the fusion process.
The fused features are fed into the reconstruction process to acquire the SR image.

Among the four processes, the alignment process is peculiar
to burst SR.
If non-aligned features are directly used for SR, the SR image may be blurred.
In~\cite{DBLP:conf/cvpr/LuoYMLJF0L21,DBLP:conf/cvpr/Guo0MR022}, deformable convolution  (DC)~\cite{DBLP:conf/iccv/DaiQXLZHW17} is used for implicit spatial alignment.
The two-step alignment with optical flow estimation and DC is also proposed in~\cite{DBLP:conf/cvpr/LuoLCYWWFSL22}.

The reconstruction process is also essential to avoid blurry images.
However, all burst SR methods introduced in Sec.~\ref{subsection: related_burst_SR} reconstruct SR images in a deterministic manner, which is prone to be blurry SR images.

\subsection{SISR with Diffusion Models}
\label{subsection: related_SISR_diffusion}

The stochasticity of diffusion models allows us to reduce blur in reconstructed images.
As with other image enhancement and restoration tasks~\cite{DBLP:conf/cvpr/ChungKKY23,DBLP:conf/icml/MurataSLTUME23,DBLP:conf/iclr/ChungKMKY23,DBLP:conf/nips/ChungSRY22,DBLP:conf/nips/KawarEES22,DBLP:journals/corr/abs-2211-12343}, an input degraded image  (i.e., LR image in the SR task) is used for conditioning the diffusion model for SISR.
For this conditioning, in SR3~\cite{DBLP:journals/pami/SahariaHCSFN23}, the LR image is upscaled by Bicubic interpolation and concatenated with images passing through diffusion steps.
In LDMs~\cite{DBLP:conf/cvpr/RombachBLEO22}, features extracted from the LR image are fed into the middle layers of U-Net in the reverse process through the cross attention mechanism.
As well as the features extracted from the LR image, the number of the diffusion steps is also used for step-aware conditioning in Stable SR~\cite{DBLP:journals/corr/abs-2305-07015}.
ResShift~\cite{DBLP:journals/pami/YueWL25} employs the residuals between high-quality and low-quality images to balance efficiency and reconstruction quality.
CCDF~\cite{DBLP:conf/cvpr/ChungSY22} further reduces the number of diffusion steps by better initialization than simple upscaling.

Unlike these SISR methods~\cite{DBLP:journals/pami/SahariaHCSFN23,DBLP:conf/cvpr/RombachBLEO22,DBLP:journals/corr/abs-2305-07015,DBLP:journals/pami/YueWL25,DBLP:conf/cvpr/ChungSY22}, this paper proposes the diffusion model conditioned with burst LR features.

\subsection{Burst Super-Resolution with Diffusion Models}
\label{subsection: related_bsr_diffusion}

For multi-frame SR, such as video SR~\cite{DBLP:conf/cvpr/NahTGBHMSL19,DBLP:conf/cvpr/HarisSU19,DBLP:conf/eccv/FuoliHGTREKXLXW20,DBLP:conf/cvpr/HarisSU20} and burst SR, spatial alignment between frames is essential.
Many burst SR methods~\cite{DBLP:conf/cvpr/BhatDTCCCCCDFGG22,DBLP:journals/corr/abs-2106-03839,DBLP:conf/cvpr/DudhaneZ0K023,DBLP:conf/cvpr/DudhaneZ0K022,DBLP:conf/cvpr/MehtaDMZ0K22} implement the alignment process in the feature domain because of its superiority compared to the image domain.
In addition, the effectiveness of the hierarchical alignment is validated in many vision tasks~\cite{DBLP:conf/iccv/LiuL00W0LG21,DBLP:conf/ijcai/ZhouNLSL20,DBLP:conf/cvpr/XuSYMY22,DBLP:conf/cvpr/LiuN0W00022}.
For burst SR using diffusion models also, BSRD~\cite{DBLP:conf/ijcnn/TokoroAU24} employs the hierarchical feature alignment process.
As one of such hierarchical feature alignment and fusion frameworks, Burstormer~\cite{DBLP:conf/cvpr/DudhaneZ0K023} is employed in BSRD.

\paragraph{Burst Feature Conditioning for Reconstruction with Diffusion Model}

In BSRD, the feature map obtained from LR images, as described above and indicated by ``Burst Image Encoder'' in Fig.~\ref{fig: teaser}, is used to condition the diffusion model to reconstruct the SR image that fits with burst LR images.
Since BSRD is implemented with U-Net, the feature map is rescaled to the $xy$ dimensions of these hierarchical layers by Bicubic interpolation to smoothly condition all hierarchical layers in U-Net.
This conditioning is achieved through Spatial Feature Transformation~\cite{DBLP:conf/cvpr/WangYDL18}.

The aforementioned conditioning process is performed in all steps in the reverse process.
This reverse process is regarded as the reconstruction process in BSRD.

\paragraph{Efficient and High-quality SR Reconstruction by the Reverse Process from Intermediate Steps}

\begin{figure}[t]
  \begin{center}
     \includegraphics[width=\linewidth]{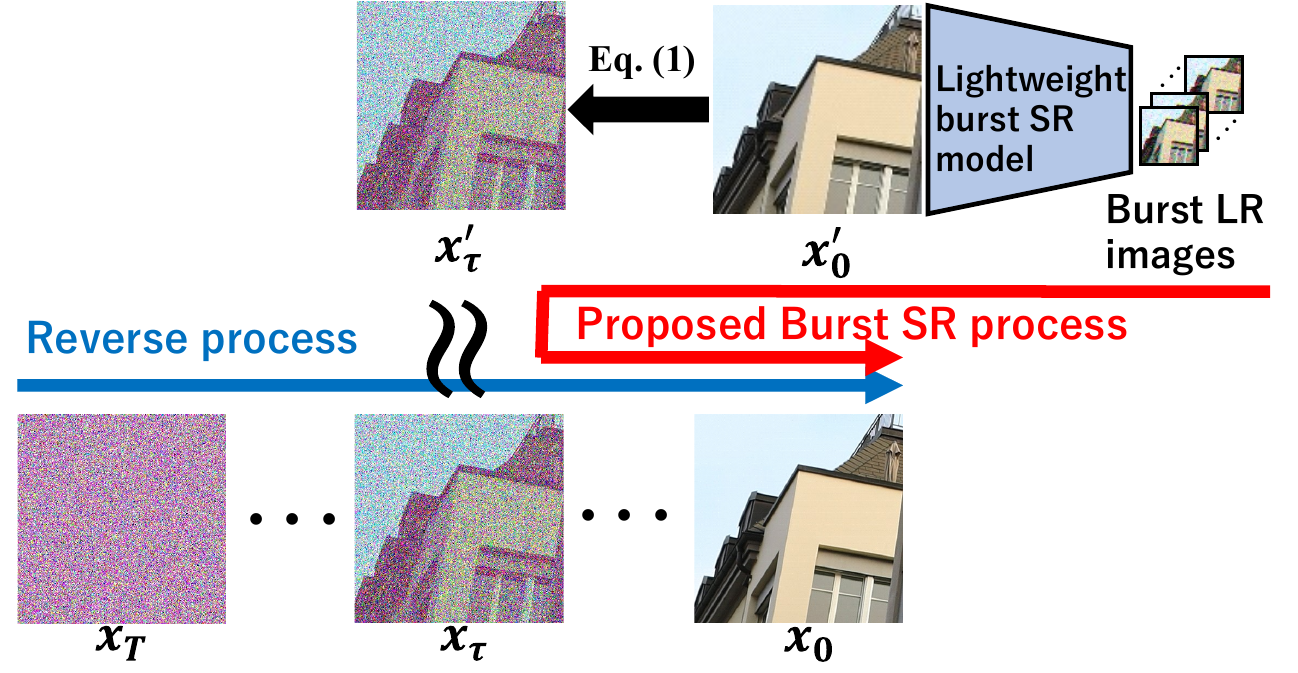}
     \vspace*{-3mm}
     \caption{Reverse process from the intermediate step for the early-step skip.
     In BSRD, the initial burst SR image $\bm{x}'_{0}$ is appropriately noised by Eq.~ (\ref{equation: initial_bsr}) and fed into the diffusion model from $\tau$-th step to skip the early diffusion steps between $T$ and $\tau$.}
     \label{fig: BSRD_reverse}
   \end{center}
   \vspace*{-3mm}
\end{figure}

To suppress the computational cost of the diffusion model, BSRD takes an initial burst SR image instead of random noise.
To start the reverse process from the initial burst SR image, it is fed into an intermediate step, instead of $T$-th step, following SDEdit~\cite{DBLP:conf/iclr/MengHSSWZE22}.
This intermediate step is denoted by $\tau$.
With this reverse process from the intermediate step, the number of execution steps is reduced, resulting in reducing computational costs.
Furthermore, 
the reverse process can be trained only between $\tau$-th and $1$st steps to focus on fine details that are important for the SR task.

The aforementioned reverse process is illustrated in Fig.~\ref{fig: BSRD_reverse}.
$x_{t}$ denotes the image at $t$-th step in the diffusion model.
In BSRD, $x_{\tau}$ is replaced by $x_{\tau}'$ generated from the initial burst SR image denoted by $x_{0}'$.
$x_{0}'$ can be provided by any burst SR method, ``Lightweight burst SR model'' in Fig.~\ref{fig: BSRD_reverse}.
While $x_{0}'$ corresponds to $x_{0}$ with no diffusion noise, $x_{0}'$ may be more blurred than $x_{0}$ because $x_{0}'$ is reconstructed in a deterministic manner.
Furthermore, $x_{\tau}'$ computed from $x_{0}'$ must contain the diffusion noise included in $x_{\tau}$ in order to replace $x_{\tau}$ with $x_{\tau}'$.
We assume that $x_{\tau}$ can be approximated by $x_{\tau}'$ by giving the diffusion noise at $\tau$-th step to $x_{0}'$.
That is, small differences caused by the blur between $x_{0}$ and $x_{0}'$ can be drowned out by the diffusion noise if $\tau$-th step is sufficiently apart from $0$-th step.
$x_{\tau}'$ is computed from $x_{0}'$ as follows:
\begin{eqnarray}
    x_{\tau}' &=& \sqrt{\smash[b]{\mathstrut\bar{\alpha}_{\tau}}}x_0' + \sqrt{\smash[b]{1 - \bar{\alpha}_{\tau}}}\epsilon, 
    \label{equation: initial_bsr} \\
    \bar{\alpha}_{\tau} &=& \prod_{s=1}^{\tau}\alpha_s,
    \label{equation: ddpm_alpha} \\
    \alpha_{t} &=& 1 - \beta_{t},
\end{eqnarray}
where $\epsilon$ denotes the zero-mean Gaussian noise with $\sigma=1$, and $\beta_{t}$ is a constant between 0 and 1 that controls the noise level at $t$-th step.



\begin{figure}[t]
  \begin{center}
     \includegraphics[width=\linewidth]{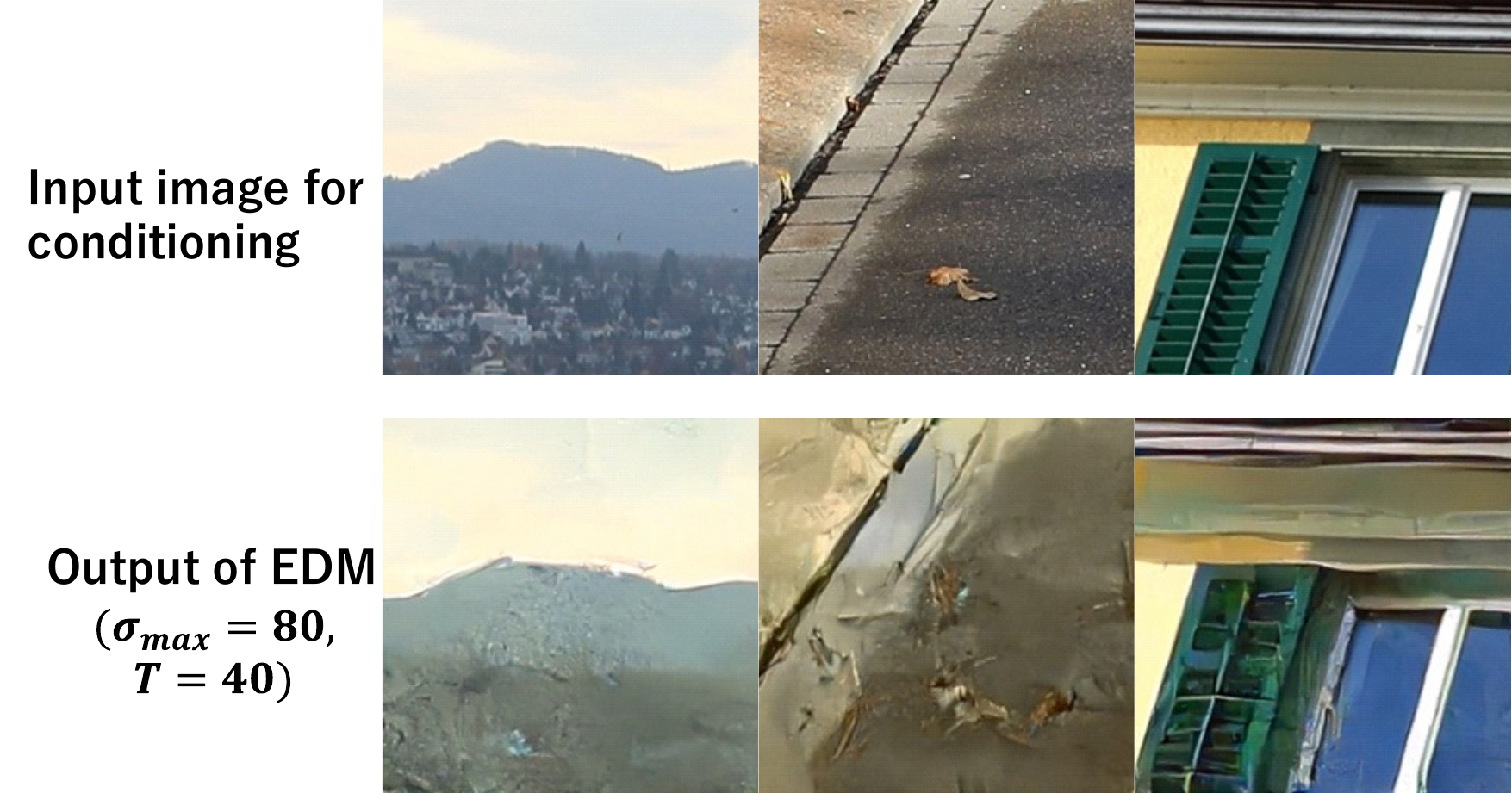}
     \caption{Examples reconstructed by EDM. Images in the lower are reconstructed by EDM conditioned by those in the upper.}
     \label{fig: EDM_results}
   \end{center}
\end{figure}

\begin{figure*}[t]
  \begin{center}
     \includegraphics[width=\textwidth]{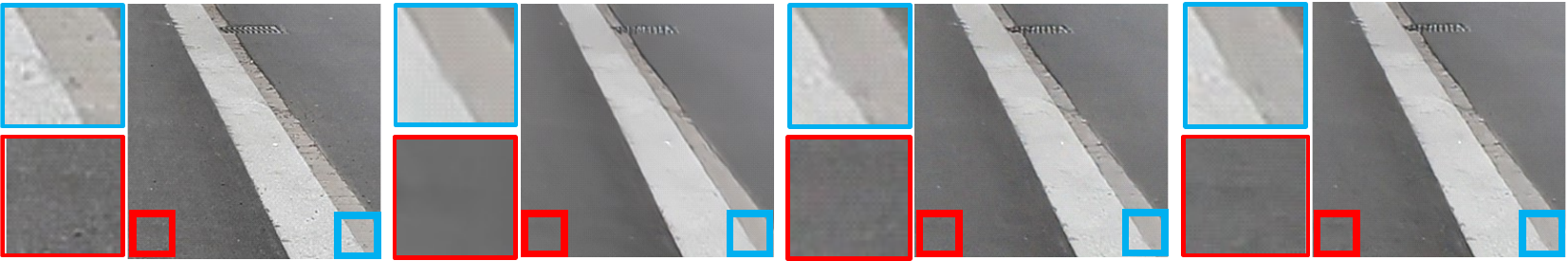}
     (a) Ground truth \hspace*{26mm}
     (b) $\tau=1$ \hspace*{26mm}
     (c) $\tau=5$ \hspace*{26mm}
     (d) $\tau=40$
     \caption{Examples of SR images reconstructed with different $\tau$ in BSRD using EDM on the BurstSR dataset.}
     \label{fig: edm_steps_visual}
  \end{center}
  \vspace*{-4mm} 
\end{figure*}

\begin{figure}[t]
    \centering
    \includegraphics[width=\linewidth]{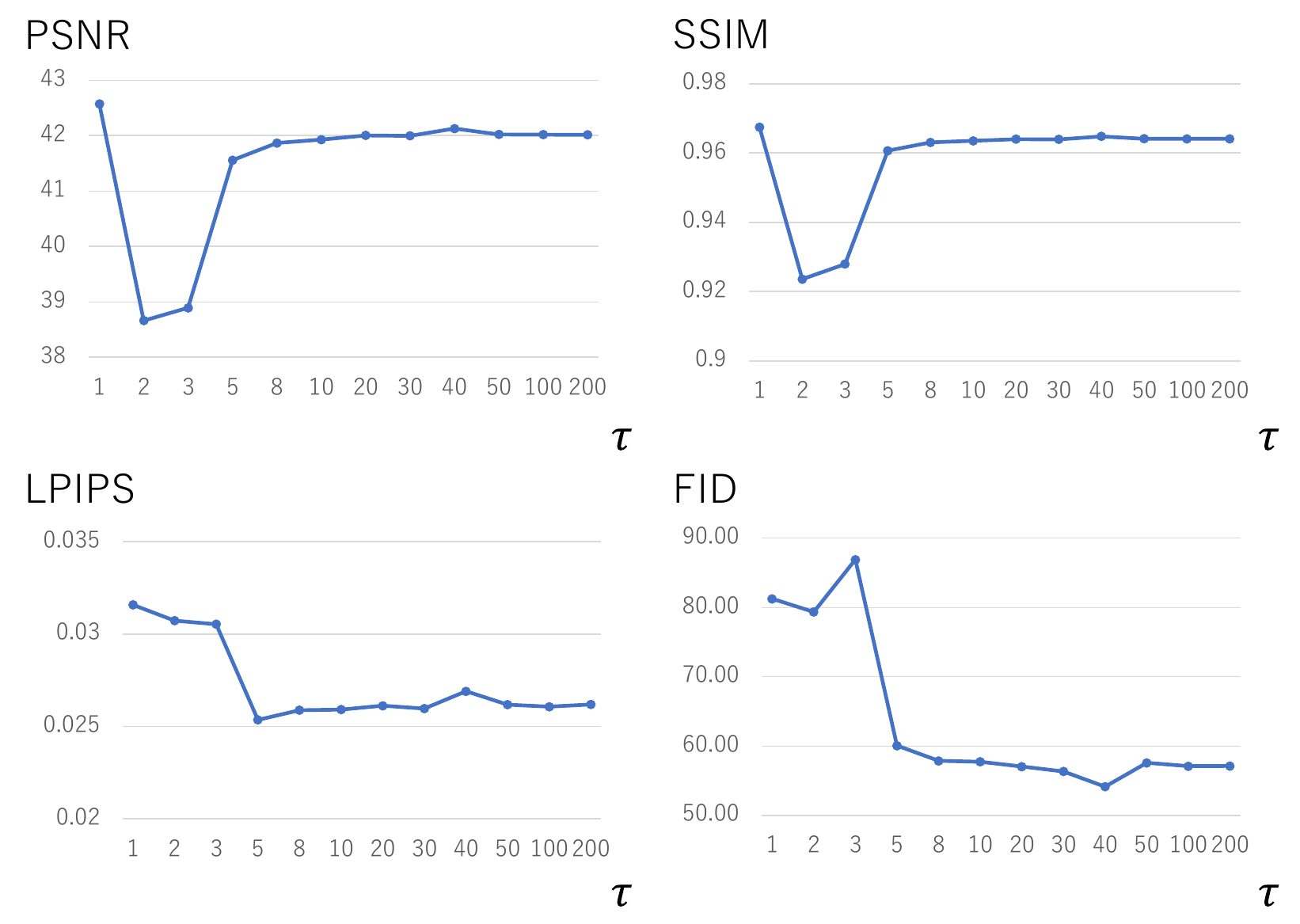}
    \caption{Effect of the number of diffusion steps, $\tau$, in BSRD using EDM on the BurstSR dataset.
    The vertical and horizontal axes indicate the metric scores and $\tau$ (logarithmic scale).
    }
    \label{fig: edm_step_effect}
\end{figure}


\section{E-BSRD: Efficient BSRD with One-step Diffusion}
\label{section: one-step}

While BSRD introduced in Sec.~\ref{subsection: related_bsr_diffusion} reduces diffusion steps used for SR reconstruction while improving the SR quality, in particular, the perceptual quality, its computational cost is still high for end users.
To further reduce the diffusion steps to one step in our proposed method based on BSRD, Sec.~\ref{section: one-step} proposes to integrate the two schemes for diffusion-step reduction, introduced in Sec.~\ref{subsection: step_reduction}.
In our proposed method described in Sec.~\ref{subsection: one-step}, these two schemes are optimized for BSRD so that their advantages compensate for the other's disadvantages for efficient one-step diffusion.

\subsection{Diffusion-step Reduction Schemes}
\label{subsection: step_reduction}

\paragraph{EDM: High-order ODE-based Stochastic Sampler}

The first scheme for diffusion-step reduction is an improved sampling process.
Our choice is EDM~\cite{DBLP:conf/nips/KarrasAAL22}, in which a higher-order differential equation is used.
EDM uses a second-order solver, which provides a good tradeoff between accuracy and efficiency, while many prior methods rely on Euler’s method.
In EDM, the preconditioning and training processes are also tweaked.
EDM reduces the diffusion steps while maintaining data generation fidelity.

However, since EDM is tweaked for image generation from random noise, it is not appropriate to directly apply it to image restoration and enhancement, including burst SR, with image conditioning.
Several examples reconstructed from such random noise are shown in Fig.~\ref{fig: EDM_results}.
The images used for conditioning (shown in the upper of Fig.~\ref{fig: EDM_results}) and the image reconstructed with these conditioning inputs are different in terms of color and textures.
The differences in color and textures are caused mainly during the early and late diffusion steps in the reverse process, respectively, as validated in~\cite{DBLP:conf/cvpr/ChoiLSKKY22}.

\paragraph{CM: One-step Diffusion with Knowledge Distillation}

The second scheme for diffusion-step reduction is to reduce the maximum diffusion steps by distilling a large diffusion model as a teacher model into a small student diffusion model.
The distilled knowledge about how to reach clean data from noisy data allows us to directly map noise to clean data without iterative diffusion steps, as proposed in the consistency model~\cite{DBLP:conf/icml/SongD0S23}, CM.

However, the image generation quality of CM is clearly degraded from its original teacher model if image generation is achieved from random noise, as verified in~\cite{DBLP:conf/icml/SongD0S23}.
In~\cite{DBLP:conf/icml/SongD0S23}, an iterative process is also proposed for further improving the data generation quality, it is still insufficient to maintain the quality if the number of iterative steps (denoted by $T_{CM}$) is smaller.

\subsection{E-BSRD: One-step Diffusion for Efficient Burst SR}
\label{subsection: one-step}

\paragraph{Initial Burst SR Image for EDM}

While EDM can reduce diffusion steps, denoising from random noise is insufficient regarding SR quality, as validated in Fig.~\ref{fig: EDM_results}.
In addition, denoising from random noise is inefficient for image restoration and enhancement tasks such as SR, as mentioned in Sec.~\ref{subsection: related_bsr_diffusion}.
Therefore, as with DDPM used in BSRD, EDM begins the reverse process from an initial burst SR image by the early-step skipping process in our proposed method.

While many parameters in the preconditioning and training processes are tweaked in EDM, we empirically found that the most important one for the early-step skipping process used in BSRD (which is shown in Fig.~\ref{fig: BSRD_reverse}) is the maximum amount of noise, $\sigma_{\text{max}}$.
This is because the default value of $\sigma_{\text{max}}$ is proposed in the original EDM paper not for the early-step skipping process using an initial image but for data generation from random noise.
The effect of $\sigma_{\text{max}}$ is empirically verified later (in Table~\ref{table: synth_sigma_max} and Table~\ref{table: real_sigma_max}).

In addition to $\sigma_{\text{max}}$, the number of diffusion steps is also crucial.
$T = 1000$ and $\tau = 100$ are different for DDPM used in BSRD~\cite{DBLP:conf/ijcnn/TokoroAU24} because BSRD pretrains the diffusion model with $T$ steps and then finetune it only from $\tau$-th step to 1st step.
For EDM, on the other hand, this pretraining may not be required because of the better convergence ability of EDM.

In our preliminary experiments, $\tau=40$ diffusion steps of EDM allow us to maintain the high burst SR quality: given the initial burst SR image reconstructed by Burstormer~\cite{DBLP:conf/cvpr/DudhaneZ0K023}, PSNR = 40.74 and 42.00 in ``BSRD with $T=1000$ and $\tau=100$ steps of DDPM'' and ``BSRD with $\tau=40$ steps of EDM,'' respectively, on the SyntheticBurst dataset~\cite{DBLP:conf/cvpr/BhatDGT21} (i.e., SR from $32 \times 32$ pixels to $256 \times 256$ pixels).

To further verify (1) whether or not the pretraining with $T$ steps is required for EDM and (2) if the pretraining is not required, how many steps (i.e., $\tau$) are required for sufficient SR quality with EDM, the qualitative and quantitative effects of varying $\tau$ in BSRD using EDM are verified in Fig.~\ref{fig: edm_steps_visual} and Fig.~\ref{fig: edm_step_effect}, respectively, while $T=40$ from random noise is proposed in the original EDM~\cite{DBLP:conf/nips/KarrasAAL22}.
Note that all results shown in Fig.~\ref{fig: edm_steps_visual} and Fig.~\ref{fig: edm_step_effect} are obtained with BSRD trained only with $\tau$ steps from an initial burst SR image given by Burstormer (i.e., without pretraining using $T$ steps).
In (a) the ground truth HR image of Fig.~\ref{fig: edm_steps_visual}, we can see texture patterns on the road.
In (b), on the other hand, such texture patterns almost disappear (i.e., oversmoothed).
This is because only one diffusion step is insufficient to reconstruct such texture patterns, which are not observed in the initial burst SR image oversmoothed due to training using a deterministic image reconstruction error such as L1 and L2 losses.
The texture patterns can be reconstructed as $\tau$ increases.
In Fig.~\ref{fig: edm_step_effect}, we can see that image distortion-based scores (i.e., PSNR and SSIM) significantly decrease in the second and third steps, while they increase around five diffusion steps and are almost saturated.
On the other hand, FID decreases gradually until around the 40th step.
In our experiments, therefore, the default value of $\tau$ in EDM is 40.

\paragraph{Compensating Integration of EDM and Consistency Model with Early-step Skipping}

\begin{figure}[t]
  \begin{center}
     \includegraphics[width=\linewidth]{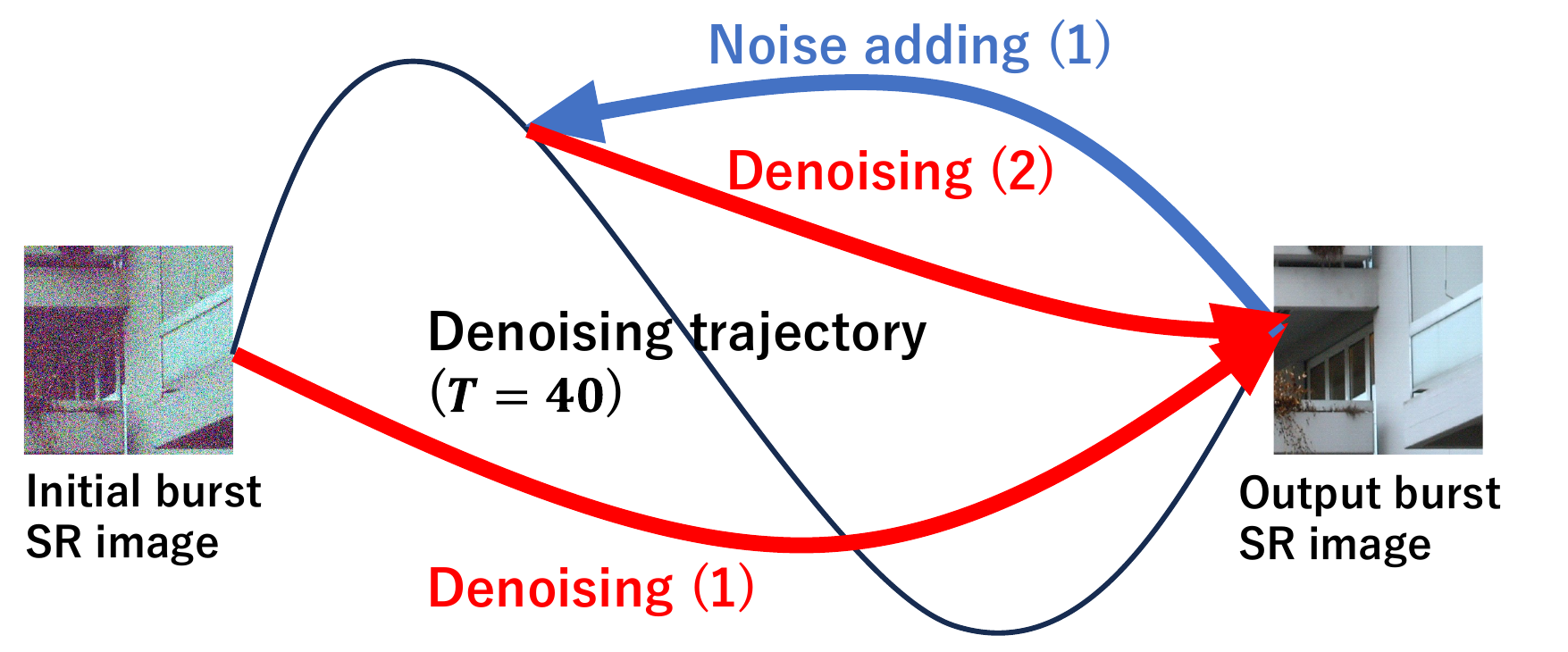}
     \caption{E-BSRD using CM. This figure shows an example with $T_{CM} = 2$. The figure in parentheses denotes the execution sequence of each process.}
     \label{fig: E-BSRD-CM}
   \end{center}
\end{figure}

While EDM can reduce the diffusion steps, the inference cost decreases only almost in proportion to the number of the diffusion steps, $\tau$.
Therefore, the reduction rate only with EDM is insufficient: 37.1 sec  (BSRD with 100 steps of DDPM) to 19.4 sec  (BSRD with 40 steps of EDM).

On the other hand, CM can further reduce the diffusion steps, while its data generation quality sometimes declines significantly if the number of the diffusion steps is reduced to one for real-time use.
Let BSRD-CM be CM distilled from BSRD modified to reconstruct the burst SR image not from an initial burst SR image but from random noise.
In our preliminary experiments, for example, the mean PSNR scores of BSRD with 100 steps and BSRD-CM are 40.74 and 27.53 on the SyntheticBurst dataset~\cite{DBLP:conf/cvpr/BhatDGT21}  (i.e., burst SR from $32 \times 32$ pixels to $256 \times 256$ pixels).

However, the aforementioned problem is avoided by distillating BSRD, which reconstructs the burst SR image not from random noise but from an initial burst SR image.
This BSRD is used as the teacher model for CM to achieve one-step diffusion.
The diffusion model distilled from BSRD by using CM is called Efficient BSRD, E-BSRD in short.

E-BSRD using CM is illustrated in Fig.~\ref{fig: E-BSRD-CM}.
CM allows for iterative steps to further improve the data generation quality, as introduced in Sec.~\ref{subsection: step_reduction}.
The number of the iterative steps is $T_{CM} = 2$ in Fig.~\ref{fig: E-BSRD-CM}.
By iteratively reducing the uncertainty of the result of CM, the data generation quality can be improved.


\begin{table}[t]
\centering
\caption{Comparison of runtime represented by secs/image.}
\label{table: Time_Comparison}
\begin{tabular}{l||l|r}
\hline
Method& Steps & Runtime\\ \hline \hline
Burstormer & NA & 0.08 \\ \hline
BIPNet & NA & 0.40\\ \hline
BSRD  & $\tau=100$ & 27.2\\ \hline
E-BSRD w/o CM & $\tau=40$ & 9.00\\ \hline
E-BSRD    & $T_{CM}=1$& 0.44\\ \hline
E-BSRD    & $T_{CM}=3$& 0.69\\ \hline
E-BSRD    & $T_{CM}=10$& 1.80\\ \hline    
\end{tabular}
\end{table}

\begin{table}[t]
\centering
\caption{Comparison of burst SR images reconstructed with varying $\sigma_{\text{max}}$ on the SyntheticBurst dataset.
``+BS'' and ``+BIP'' mean that initial burst SR images are given by Burstormer and BIPNet, respectively.}
\label{table: synth_sigma_max}
\resizebox{85mm}{!}{
\begin{tabular}{l|r||rrrr}
\hline
Method& $\sigma_{\text{max}}$ & PSNR$\uparrow$& SSIM$\uparrow$& LPIPS$\downarrow$& FID$\downarrow$\\ \hline \hline
Burstormer& NA & \red{43.21}& \red{0.971}& 0.030& 70.76\\ \hline
BSRD  (+BS) & NA & 40.74 & 0.952 & \blue{0.027} & \red{50.61}\\ \hline
E-BSRD  (+BS) & 80                           & 32.91 & 0.863 & 0.101 & 176.6\\ \hline
E-BSRD  (+BS) & 0.2& 38.66 & 0.933 & 0.040 & 110.3\\ \hline
E-BSRD  (+BS) & 0.08& 40.66 & 0.954 & 0.030 & 83.03 \\ \hline
E-BSRD  (+BS) & 0.05& 41.58& 0.961&  0.028& 68.94\\ \hline
E-BSRD  (+BS) & 0.03& 42.00& 0.964& \red{0.026}& 56.71\\ \hline
E-BSRD  (+BS) & 0.01& 42.68& 0.968& 0.028& \blue{51.90}\\ \hline
E-BSRD  (+BS) & 0.005& \blue{42.95}& \blue{0.970}& 0.029& 59.05\\ \hline \hline
BIPNet&  NA & \red{42.61} & \red{0.968} & 0.032 & 64.17 \\ \hline
BSRD  (+BIP) & NA & 40.53 & 0.951 & \blue{0.029} & \red{50.26}\\ \hline
E-BSRD  (+BIP)     & 80& 32.93& 0.861& 0.102& 173.9\\ \hline
E-BSRD  (+BIP)     & 0.2& 38.47& 0.932& 0.042& 112.7\\ \hline
E-BSRD  (+BIP)     & 0.08& 40.37& 0.952& 0.032& 84.65\\ \hline
E-BSRD  (+BIP)     & 0.05& 41.22& 0.959& 0.030& 68.43\\ \hline
E-BSRD  (+BIP)     & 0.03& 41.58& 0.962& \red{0.028}& 56.41\\ \hline
E-BSRD  (+BIP)     & 0.01& 42.17& 0.966& 0.030&  \blue{50.75}\\ \hline
E-BSRD  (+BIP)     & 0.005& \blue{42.39}& \blue{0.967}& 0.031& 56.24\\ \hline
\end{tabular}
}
\vspace*{4mm}
\centering
\caption{Comparison between E-BSRD w/o CM and E-BSRD w/ CM on the SyntheticBurst dataset.}
\resizebox{85mm}{!}{
\label{table: synth_CM_Comparison_with_EBSRD}
\begin{tabular}{l |r||c c c c}
\hline
Methods&Steps& PSNR$\uparrow$& SSIM$\uparrow$& LPIPS$\downarrow$& FID$\downarrow$\\ \hline \hline
\multirow{2}{22mm}{E-BSRD w/o CM  (+BS)} & \multirow{2}{*}{$\tau=40$}& \multirow{2}{*}{42.00}& \multirow{2}{*}{\red{0.964}}& \multirow{2}{*}{\red{0.026}}& \multirow{2}{*}{\red{56.71}}\\ &&&&& \\ \hline
E-BSRD  (+BS) & $T_{CM}=1$& 42.00& \red{0.964}& \red{0.026} & 57.39\\ \hline
E-BSRD  (+BS) & $T_{CM}=2$& \red{42.02} & \red{0.964} & \red{0.026} &58.36\\ \hline
E-BSRD  (+BS) & $T_{CM}=3$& \blue{42.01}& \red{0.964}& \red{0.026}& \blue{57.32}\\ \hline
E-BSRD  (+BS) & $T_{CM}=4$&  41.99 & \red{0.964} &  \red{0.026} & 58.13\\ \hline
E-BSRD  (+BS) & $T_{CM}=5$&  41.89 &  0.963 & \red{0.026} & 60.03\\ \hline
E-BSRD  (+BS) & $T_{CM}=11$&  41.65 &  \blue{0.962} &  \red{0.026} & 62.47\\ \hline \hline
\multirow{2}{22mm}{E-BSRD w/o CM  (+BIP)}& \multirow{2}{*}{$\tau=40$}& \multirow{2}{*}{\blue{41.58}}& \multirow{2}{*}{\red{0.962}}& \multirow{2}{*}{\red{0.028}}& \multirow{2}{*}{\red{56.41}} \\
&&&&& \\ \hline
E-BSRD  (+BIP)& $T_{CM}=1$ & \blue{41.58}& \red{0.962}& \red{0.028}& 58.21\\ \hline
E-BSRD  (+BIP)& $T_{CM}=2$& \red{41.59} & \red{0.962} & \red{0.028} & \blue{58.06}\\ \hline
E-BSRD  (+BIP)& $T_{CM}=3$& \blue{41.58}& \red{0.962}&  \red{0.028} & 58.07\\ \hline
E-BSRD  (+BIP)& $T_{CM}=4$&  41.56 &  \red{0.962} &  \red{0.028} & 59.08\\ \hline
E-BSRD  (+BIP)& $T_{CM}=5$&  41.48 &  0.961 &  \red{0.028} & 59.44\\ \hline
E-BSRD  (+BIP)& $T_{CM}=11$ &  41.26 &  0.959 &  \red{0.028} & 63.18\\ \hline
\end{tabular}
}
\end{table}

\begin{table}[t]
\centering
\caption{Comparison of burst SR images reconstructed with varying $\sigma_{\text{max}}$ on the BurstSR dataset.}
\label{table: real_sigma_max}
\resizebox{85mm}{!}{%
\begin{tabular}{l|r||cccc}
\hline
Methods & $\sigma_{\text{max}}$& PSNR$\uparrow$& SSIM$\uparrow$& LPIPS$\downarrow$&FID$\downarrow$\\ \hline \hline
Burstormer & NA & \red{50.49} & \red{0.985}  & 0.056& 74.21 \\ \hline
BSRD (+BS) & NA & 49.45 & 0.915 & \red{0.050} & \red{48.51} \\ \hline
E-BSRD (+BS) & 0.1& 48.36 & 0.979 & 0.057 & 85.83 \\ \hline
E-BSRD (+BS) & 0.08 & 48.58 & 0.908 & 0.056 & 81.95 \\ \hline
E-BSRD (+BS) & 0.05 & 49.21 & 0.982 & \blue{0.055} & 78.92 \\ \hline
E-BSRD (+BS) & 0.03 & 49.21 & 0.982 & \blue{0.055} & {63.70} \\ \hline
E-BSRD (+BS) & 0.01 & 49.88 & \blue{0.984} & \blue{0.055} & 65.03 \\ \hline
E-BSRD (+BS) & 0.005 & \blue{50.14} & \blue{0.984} & \blue{0.055} & \blue{62.49} \\ \hline \hline
BIPNet  & NA &  \red{51.55} &  \red{0.986}  & 0.050 & 75.43 \\ \hline
BSRD (+BIP) & NA&50.54 & 0.984 &  \red{0.047} &  \red{43.34} \\ \hline
E-BSRD (+BIP) & 0.1 & 48.94 & 0.980 & 0.050 & 82.10 \\ \hline
E-BSRD (+BIP) & 0.08 & 49.17 & 0.981 & 0.050 & 81.95 \\ \hline
E-BSRD (+BIP) & 0.05 & 49.97 & 0.983 & 0.049 & 73.82 \\ \hline
E-BSRD (+BIP) & 0.03 & 49.96 & 0.984 & \blue{0.048} & 60.37 \\ \hline
E-BSRD (+BIP) & 0.01 & 50.73   & \blue{0.985} & \blue{0.048} & \blue{57.28} \\ \hline
E-BSRD (+BIP) & 0.005 & \blue{51.07} & \blue{0.985} & 0.049 & 58.00 \\ \hline
\end{tabular}
}
\vspace*{4mm}
\centering
\caption{Comparison between E-BSRD w/o CM and E-BSRD w/ CM on the BurstSR dataset.}
\resizebox{85mm}{!}{%
\label{table: real_CM_Comparison_with_DSRED}
\begin{tabular}{l|r||cccc}
\hline
Methods & Steps &PSNR$\uparrow$ & SSIM$\uparrow$ & LPIPS$\downarrow$ & FID$\downarrow$                          \\ \hline \hline
\multirow{2}{22mm}{E-BSRD w/o CM (+BS)} & \multirow{2}{*}{$\tau=40$} &  \multirow{2}{*}{\red{49.21}} &  \multirow{2}{*}{\red{0.982}} & \multirow{2}{*}{\red{0.055}} &  \multirow{2}{*}{\red{63.70}} \\
&&&&& \\ \hline
E-BSRD (+BS)   & $T_{CM}=1$ & 47.64 & \blue{0.979} & 0.060 & 77.30 \\ \hline
E-BSRD (+BS)   & $T_{CM}=2$ & 47.66 & 0.978 & 0.060 & 77.40 \\ \hline
E-BSRD (+BS)   & $T_{CM}=3$ & 47.64 & \blue{0.979} & 0.060 & 77.01 \\ \hline
E-BSRD (+BS)   & $T_{CM}=4$ & 48.05 & \blue{0.979} & 0.058 & \blue{76.68} \\ \hline
E-BSRD (+BS)   & $T_{CM}=5$ & \blue{48.09} & \blue{0.979} & \blue{0.057} & 77.00 \\ \hline
E-BSRD (+BS)   & $T_{CM}=11$ & 47.38 & 0.976 & 0.058 & 79.47 \\ \hline \hline
\multirow{2}{22mm}{E-BSRD w/o CM (+BIP)} & \multirow{2}{*}{$\tau=40$} &  \multirow{2}{*}{\red{49.96}} &  \multirow{2}{*}{\red{0.984}} &  \multirow{2}{*}{\red{0.048}}  &  \multirow{2}{*}{\red{60.37}} \\
&&&&& \\ \hline
E-BSRD (+BIP) & $T_{CM}=1$ & 48.29 & 0.980 & 0.054 & 71.80 \\ \hline
E-BSRD (+BIP) & $T_{CM}=2$ & 48.31 & 0.980 & 0.053 & 72.59 \\ \hline
E-BSRD (+BIP) & $T_{CM}=3$ & 48.29 & 0.980 & 0.054 & 71.37 \\ \hline
E-BSRD (+BIP) & $T_{CM}=4$ & 48.77 & \blue{0.981} & 0.051 & \blue{70.43} \\ \hline
E-BSRD (+BIP) & $T_{CM}=5$ & \blue{48.78} & \blue{0.981} & \blue{0.050} & 72.01 \\ \hline
E-BSRD (+BIP) & $T_{CM}=11$ & 48.03 & 0.978 & \blue{0.050} & 75.26 \\ \hline
\end{tabular}
}
\end{table}

\section{Experiments}
\label{section: experiments}

\subsection{Details}

\paragraph{Evaluation Metrics}

As standard evaluation metrics, LPIPS~\cite{DBLP:conf/cvpr/ZhangIESW18}, FID~\cite{DBLP:conf/nips/HeuselRUNH17}, PSNR, and SSIM~\cite{DBLP:journals/tip/WangBSS04} are used.
While PSNR and SSIM are for image distortion-based evaluation, LPIPS and FID are proposed for perceptual quality evaluation.
Lower scores are better in LPIPS and FID, while higher scores are better in PSNR and SSIM.

\paragraph{Training details and Parameters}

We follow all training details and parameters of the original EDM~\cite{DBLP:conf/nips/KarrasAAL22}.

\paragraph{Datasets}

The SyntheticBurst and BurstSR datasets are used~\cite{DBLP:conf/cvpr/BhatDGT21}.

In the SyntheticBurst dataset, each sRGB HR image is degraded to its LR image.
This degradation process, including ISP, follows the one proposed in ~\cite{DBLP:conf/cvpr/BhatDGT21} as follows.
Burst LR images, except for the reference frame, are randomly  (1) translated up to 24 pixels along $x$ and $y$ axes and  (2) rotated up to one degree.
The SyntheticBurst dataset has 46,839 training images and 300 test images.

In the BurstSR dataset, both LR and HR images are real images.
The LR images are captured using a burst shot mode of Samsung Galaxy S8.
These LR images are subtly different from each other due to handshake effects.
Each HR image is captured by CANON 5D Mark IV.
This dataset has 5,405 training images and 882 test images.

In both of these two datasets, the center region of each original HR image is cropped out to $256 \times 256$ pixels.

\paragraph{Burst SR Condition}

In all experiments, the dimensions of LR and HR images are $32 \times 32 \times 4$ RAW images and $256 \times 256 \times 3$ RGB images, respectively.
The number of a set of burst LR images is eight.
For comparison, BIPNet~\cite{DBLP:conf/cvpr/DudhaneZ0K022}, Burstormer~\cite{DBLP:conf/cvpr/DudhaneZ0K023}, and BSRD~\cite{DBLP:conf/ijcnn/TokoroAU24} are evaluated.
%
The weights of all these burst SR models are trained by the authors' codes.

Initial burst SR images given to BSRD and our method (E-BSRD) are provided by BIPNet and Burstormer.

While the number of diffusion steps of EDM in E-BSRD is $\tau=40$, $T=1000$ and $\tau=100$ for DDPM in BSRD.

\begin{figure*}[t]
    \centering
    \includegraphics[width=\textwidth]{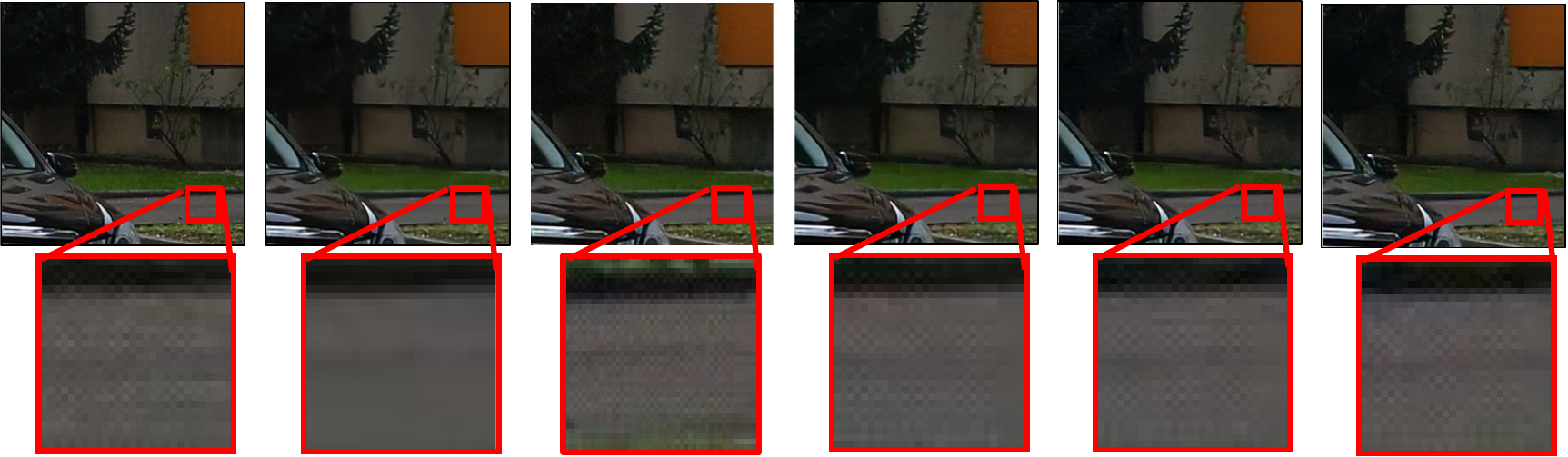}\\
    \medskip
    \includegraphics[width=\textwidth]{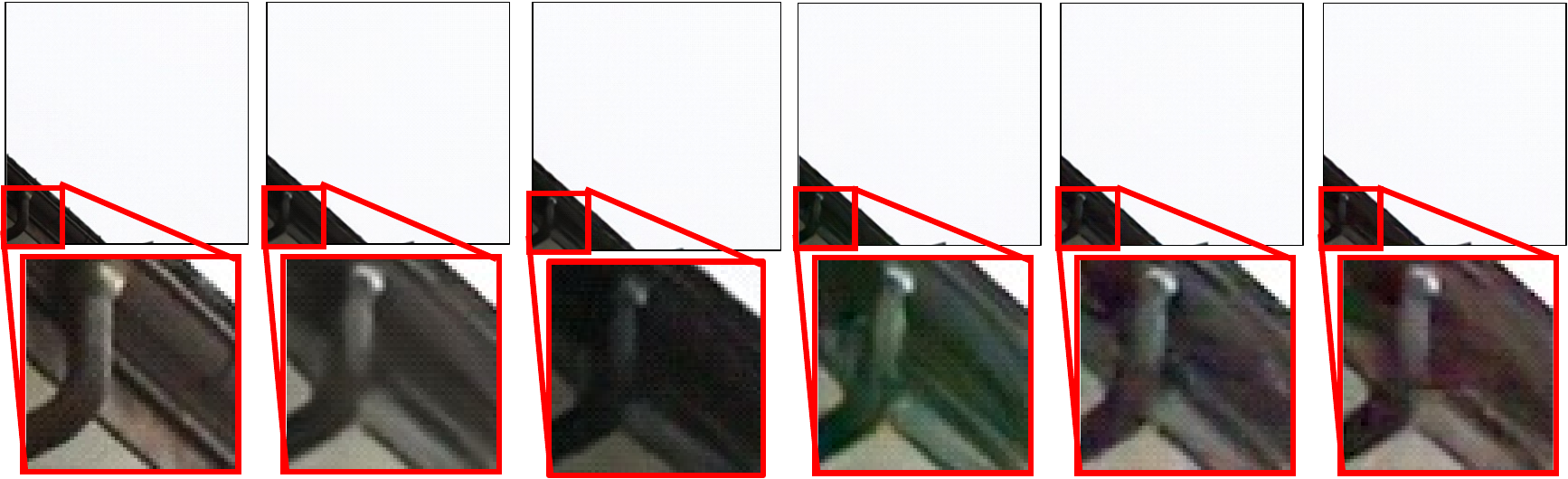}\\
    ~\hspace*{5mm}
    GT \hspace*{18mm}
    Burstormer \hspace*{16mm}
    BIPNet \hspace*{16mm}
    BSRD \hspace*{12mm}
    E-BSRD w/o CM \hspace*{8mm}
    E-BSRD
    \caption{Visual results on the SyntheticBurst dataset. Best viewed with zoom-in.
    While clear textures, such as the boundary lines of texts, are generally used as the visual results of SR, smoother and finer textures are appropriate for validating the effectiveness of our method because our method emphasizes such smooth and fine textures.}
    \label{fig: visual_comparison_synth}
    \vspace*{-4mm}
\end{figure*}

\subsection{Runtime}
\label{subsection: runtime}

Table~\ref{table: Time_Comparison} validates the efficient computational cost of E-BSRD.
As mentioned in Sec.~\ref{subsection: one-step}, the runtime decreases almost in proportion to the diffusion steps from BSRD to E-BSRD w/o CM (i.e., from 27.2 to 9.00).
In E-BSRD (i.e., E-BSRD with CM), the runtime is further significantly decreased by reducing the number of steps (i.e., $T_{CM} = \{1, 3, 10 \}$).
For example, the runtime decreases from 27.2 (BSRD with $\tau=100$) to 0.44 (E-BSRD with $T_{CM} = 1$).
That is, the runtime is reduced to 1.6 \% of BSRD.

\begin{figure*}[t]
    \centering
    \includegraphics[width=\textwidth]{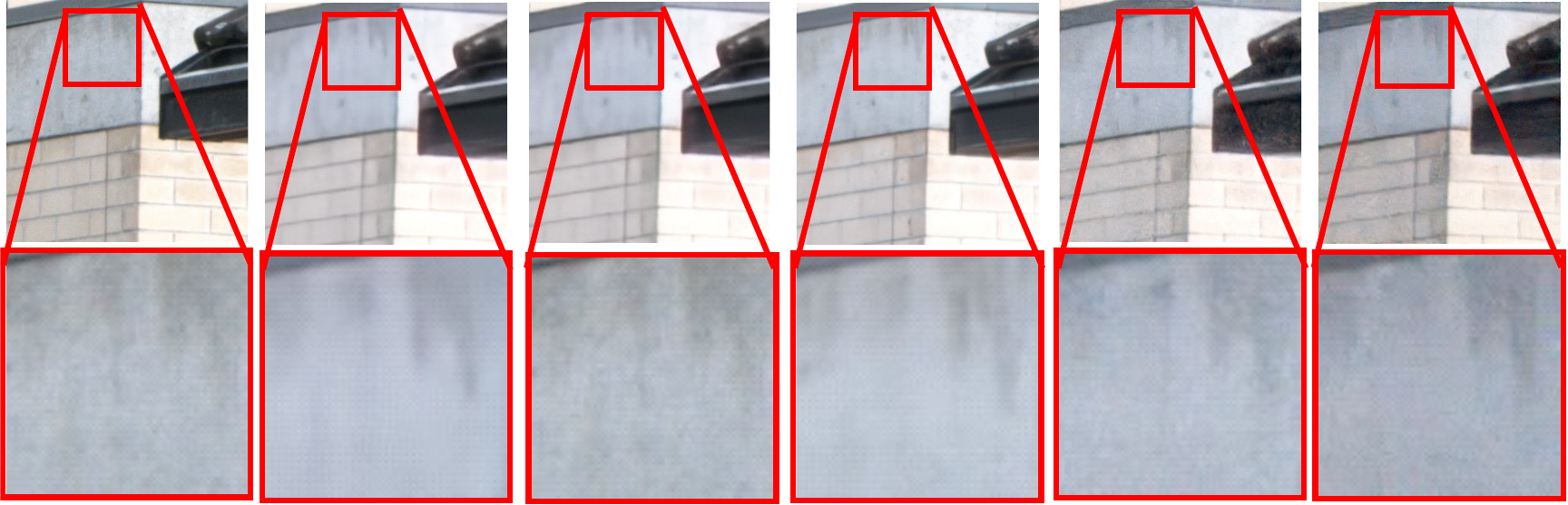}\\
    \medskip
    \includegraphics[width=\textwidth]{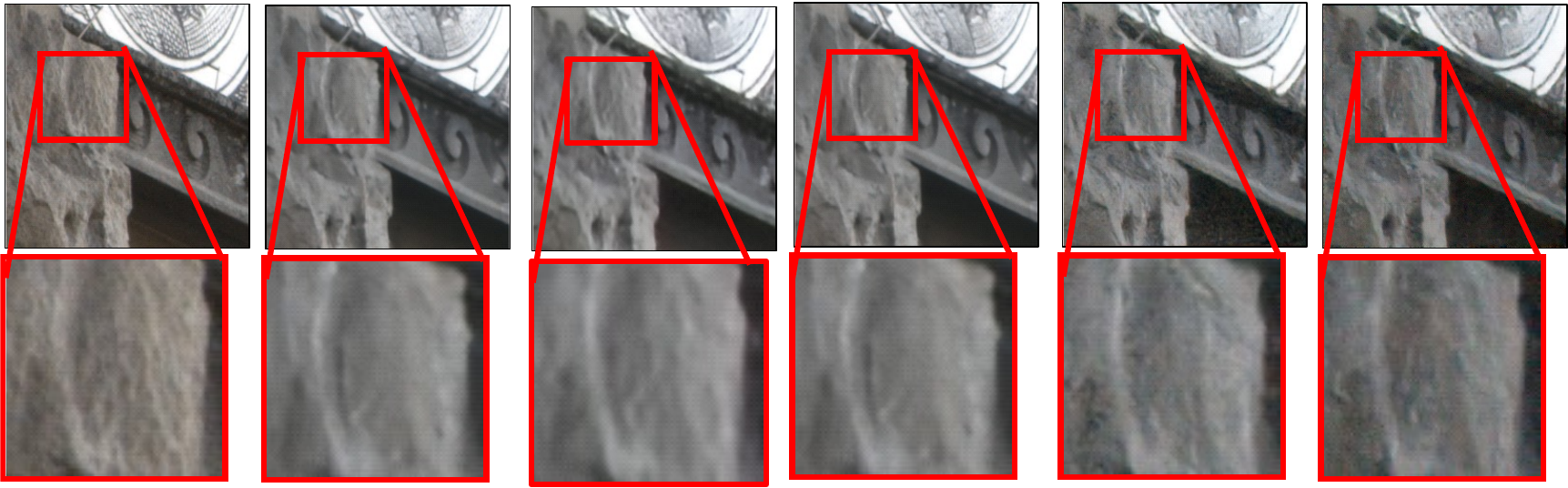}\\
    ~\hspace*{5mm}
    GT \hspace*{18mm}
    Burstormer \hspace*{16mm}
    BIPNet \hspace*{16mm}
    BSRD \hspace*{12mm}
    E-BSRD w/o CM \hspace*{8mm}
    E-BSRD
    \caption{Visual results on the BurstSR dataset. Best viewed with zoom-in.}
    \label{fig: visual_comparison_real}
    \vspace*{-4mm}
\end{figure*}


\subsection{Results: SyntheticBurst Dataset}
\label{subsection:synthetic_burst_dataset}

\paragraph{Effect of $\sigma_{\text{max}}$}

Table~\ref{table: synth_sigma_max} shows the SR scores obtained with the varying amount of noise given to the initial burst SR image.
Independent of the initial burst SR images, the SR qualities of different conditions have the same trend.
\begin{itemize}
\item The image distortion quality (i.e., PSNR and SSIM) is the highest in the original initial burst SR image reconstructed in a deterministic manner (i.e., without the diffusion model).
Both PSNR and SSIM improve as $\sigma_{\text{max}}$ decreases in our E-BSRD.
This is natural because the effect of the diffusion model is reduced as $\sigma_{\text{max}}$ decreases.
\item Compared with Burstormer, which is a deterministic model, all results obtained by diffusion models are inferior in terms of the image distortion-based scores (PSNR and SSIM). 
On the other hand, the advantage of diffusion models~\cite{DBLP:conf/ijcnn/TokoroAU24} is to improve the perceptual quality (i.e., LPIPS and FID), as validated in our experiments as well as in~\cite{DBLP:conf/ijcnn/TokoroAU24}; i.e., LPIPS and FID of BSRD are the second-best and the best, respectively.
While our E-BSRD reduces the runtime significantly as shown in Table~\ref{table: Time_Comparison}, it maintains the advantage of BSRD, i.e., the best LPIPS and the second-best FID are observed in E-BSRD.
\end{itemize}

\paragraph{Effect of Iterations in CM}

Table~\ref{table: synth_CM_Comparison_with_EBSRD} shows the SR scores obtained with the varying number of iterations of CM (i.e., $T_{CM}$).
We can see that the SR quality is almost unchanged in all metrics, except for larger $T_{CM}$, such as $T_{CM} = 11$.
Therefore, $T_{CM} = 1$ is sufficient for our E-BSRD.

Why is the data generation quality not changed by the iterations, unlike experiments in the original CM paper~\cite{DBLP:conf/icml/SongD0S23}?
This may be because the difference between the initial data (i.e., initial burst SR image) and the final output is small, while the random noise is fed into CM in the original work.

The reconstructed burst SR images are shown in Fig.~\ref{fig: visual_comparison_synth}.
For reference, the ground truth HR images are also shown.
In the upper example, Burstormer oversmoothes the SR image, as we can see in the zoom-in region.
Such oversmoothed results are shown as typical bad examples of deterministic burst SR models in both Fig.~\ref{fig: visual_comparison_synth} and Fig.~\ref{fig: visual_comparison_real}.
Another difficulty can be validated in the lower example.
It is known that the global color structure can be changed by image generation using diffusion models~\cite{DBLP:conf/cvpr/ChoiLSKKY22}.
In the results of BSRD and E-BSRD w/o CM, we can see that the global color structure is changed to greenish.
On the other hand, E-BSRD can maintain the global color structure.


\subsection{Results: BurstSR dataset}
\label{subsection:burst_sr_dataset}

Table~\ref{table: real_sigma_max} shows the quantitative results obtained by varying $\sigma_{\text{max}}$ in our E-BSRD.
The overall trend of the performance measures in Table~\ref{table: real_sigma_max} is similar to Table~\ref{table: synth_sigma_max}, which shows the results on the SyntheticBurst dataset, as follows.
The best scores in image distortion-based metrics (i.e., PSNR and SSIM) and perceptual metrics (i.e., LPIPS and FID) are acquired in the original deterministic burst SR methods (i.e., Bursotrmer and BIPNet) and BSRD, respectively.
However, the gaps from these methods and our E-BSRD are not significant, while its runtime is significantly faster than these methods, as shown in Table~\ref{table: Time_Comparison}.
For example, several best and second-best scores are observed in E-BSRD with small $\sigma_{\text{max}}$ both in E-BSRD (+BS) and E-BSRD (+BIP).

Table~\ref{table: real_CM_Comparison_with_DSRED} shows the results with varying iterative steps, $T_{CM}$, in our E-BSRD.
As with Table~\ref{table: real_sigma_max}, Table~\ref{table: real_CM_Comparison_with_DSRED} also shows the overall trend similar to Table~\ref{table: synth_CM_Comparison_with_EBSRD}, which shows the results on the SyntheticBurst dataset.
While better scores are obtained by E-BSRD without CM, the performance is not decreased if $T_{CM}$ is small in our E-BSRD with CM.

The examples of reconstructed burst SR images are shown in Fig.~\ref{fig: visual_comparison_real}.
As with the upper example shown in Fig.~\ref{fig: visual_comparison_synth}, we can see that Burstormer and BIPNet, each of which is a deterministic burst SR method, oversmoothes fine textures.
Finer textures are reconstructed by E-BSRD.



\section{Conclusion}
\label{section: conclusion}

This paper proposed a fast burst SR method using few-step diffusion.
Even if the number of diffusion steps is reduced to one, the SR quality is comparable to the method using diffusion with many steps (e.g., 100 steps)~\cite{DBLP:conf/ijcnn/TokoroAU24}.
Our proposed method, called E-BSRD, with one-step diffusion reduces the runtime to around 1.6 \% of its baseline~\cite{DBLP:conf/ijcnn/TokoroAU24}.
E-BSRD is enhanced with a high-order ODE (i.e., second-order ODE) and a teacher-student distillation approach.


This work is supported by JSPS KAKENHI 22H03618.


{
    \small
    \bibliographystyle{ieeenat_fullname}
    \bibliography{main}
}

\end{document}